\documentclass[runningheads]{llncs}
\usepackage{graphicx}
\usepackage{multirow}
\usepackage{multicol}
\usepackage[font=small,labelfont=bf]{caption}
\usepackage{url}
\usepackage[english]{babel}

\begin{document}
\title{SEMA: an Extended Semantic Evaluation Metric for AMR}
%
%
\author{Rafael T. Anchi\^{e}ta\inst{1} \and
Marco A. S. Cabezudo\inst{1} \and
Thiago A. S. Pardo\inst{1}}
\authorrunning{Anchi\^{e}ta et al.}
%
\institute{
	Interinstitutional Center for Computational Linguistics (NILC) \\
	Institute of Mathematical and Computer Sciences, University of S\~{a}o Paulo, Brazil
\email{rta@usp.br, msobrevillac@usp.br, taspardo@icmc.usp.br}}
%
\maketitle              
\begin{abstract}
Abstract Meaning Representation (AMR) is a recently designed semantic representation language intended 
to capture the meaning of a sentence, which may be represented as a single-rooted directed acyclic graph 
with labeled nodes and edges. 
The automatic evaluation of this structure plays an important role in the development 
of better systems, as well as for semantic annotation. Despite there is one available metric, 
\textit{smatch}, it has some drawbacks. For instance, \textit{smatch} creates a self-relation on the root 
of the graph, has weights for different error types, and does not take into account the dependence of the 
elements in the AMR structure. With these drawbacks, \textit{smatch} masks several problems of the 
AMR parsers and distorts the evaluation of the AMRs. In view of this, in this paper, we introduce an 
extended metric to evaluate AMR parsers, which deals with the drawbacks of the \textit{smatch} metric. 
Finally, we compare both metrics, using four well-known AMR parsers, and we argue that our metric is more 
refined, robust, fairer, and faster than \textit{smatch}.

\end{abstract}

\section{Introduction}



Abstract Meaning Representation (AMR) is a semantic representation language designed to capture the meaning 
of a whole sentence \cite{banarescu2013abstract}. AMR got the attention of the scientific community due to 
its relatively simpler structure, showing the relations among concepts and making them easy to read. 
The creation of AMR language was motivated by the need of providing to the research community corpora with 
annotations related to traditional tasks of Natural Language Processing (NLP), such as named entity 
recognition, semantic role labeling, word sense disambiguation, and coreference resolution 
\cite{banarescu2013abstract}. Moreover, AMR structures are arguably easier to produce than traditional 
formal meaning representations~\cite{bos2016expressive}.

In this way, several annotated corpora arose, for 
English\footnote{\url{https://amr.isi.edu/download.html}}, 
Chinese \cite{li2016annotating}, Spanish \cite{migueles-abraira}, and Portuguese 
\cite{anchietaEpardoAMR}. Consequently, a considerable number of semantic parsers emerged 
\cite{flanigan2014discriminative,damonte2017parser,van2017neural,anchietaPardoParser,chunchuan-amrgraph}, 
and, with the available parsers, some applications were developed and/or improved: 
automatic summarization \cite{vlachos2018guided}, text generation \cite{song18}, paraphrase detection \cite{issa2018abstract}, and others.

Given the growing interest in AMR language, the automatic evaluation of AMR structures plays a very 
important role for the AMR parsing task, as well as for semantic annotation tasks, which create 
linguistic resources for semantic parsing. Although there is one metric to automatically evaluate AMR 
structures, named \textit{smatch} \cite{cai2013smatch}, it has some shortcomings:

\begin{enumerate}
    \item \textit{Smatch} does not take into account the dependence of the elements in the AMR structure, 
    i.e., its analysis is very simple, masking several analysis problems. So, \textit{smatch} often 
    gives higher scores for AMRs that have different meanings in relation to the reference AMR.
    \item \textit{Smatch} creates a self-relation called \textit{TOP} for the root of the AMR structure. 
    That is, \textit{smatch} gives more weight for the root of the graph than other elements, distorting 
    the analysis.
    \item \textit{Smatch} has weights for different error types. As discussed by Damonte et al.
    \cite{damonte2017parser}, three named entity errors are considered 
    more important than six wrong labels. Nevertheless, it is difficult to conclude which task should have 
    a higher weight.
\end{enumerate}

\textit{Smatch} metric 
computes the degree of overlapping between two AMR structures. To evaluate an AMR generated by a parser 
against a reference manually produced AMR, \textit{smatch} defines {\tt M} the correct number of 
triples, {\tt C} the produced number of triples by a parser, and {\tt T} the total number of triples 
in reference AMR. So, precision and recall are calculated according to 
Eq.~\ref{eq:precision} and~\ref{eq:recall}, respectively.

\begin{multicols}{2}
    \begin{equation} \label{eq:precision}
        P = \frac{M}{C}
    \end{equation}\break
    \begin{equation} \label{eq:recall}
        R = \frac{M}{T}
    \end{equation}
\end{multicols}

\begin{figure}[h]
	\centering
	\begin{minipage}[b]{0.4\textwidth}
		\centering
		\includegraphics[scale=0.4]{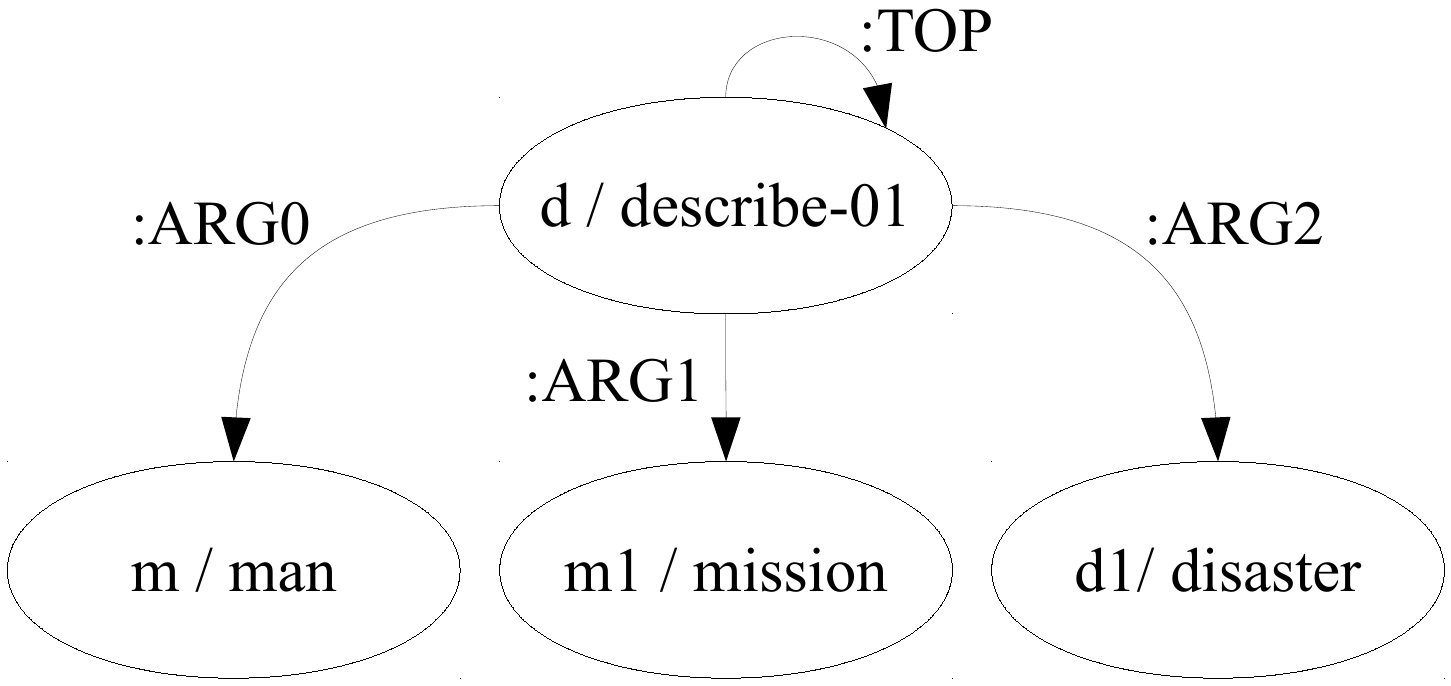}
		\caption{Reference AMR}
		\label{fig:gold}
	\end{minipage}
	\hfill
	\begin{minipage}[b]{0.47\textwidth}
		\centering
		\includegraphics[scale=0.36]{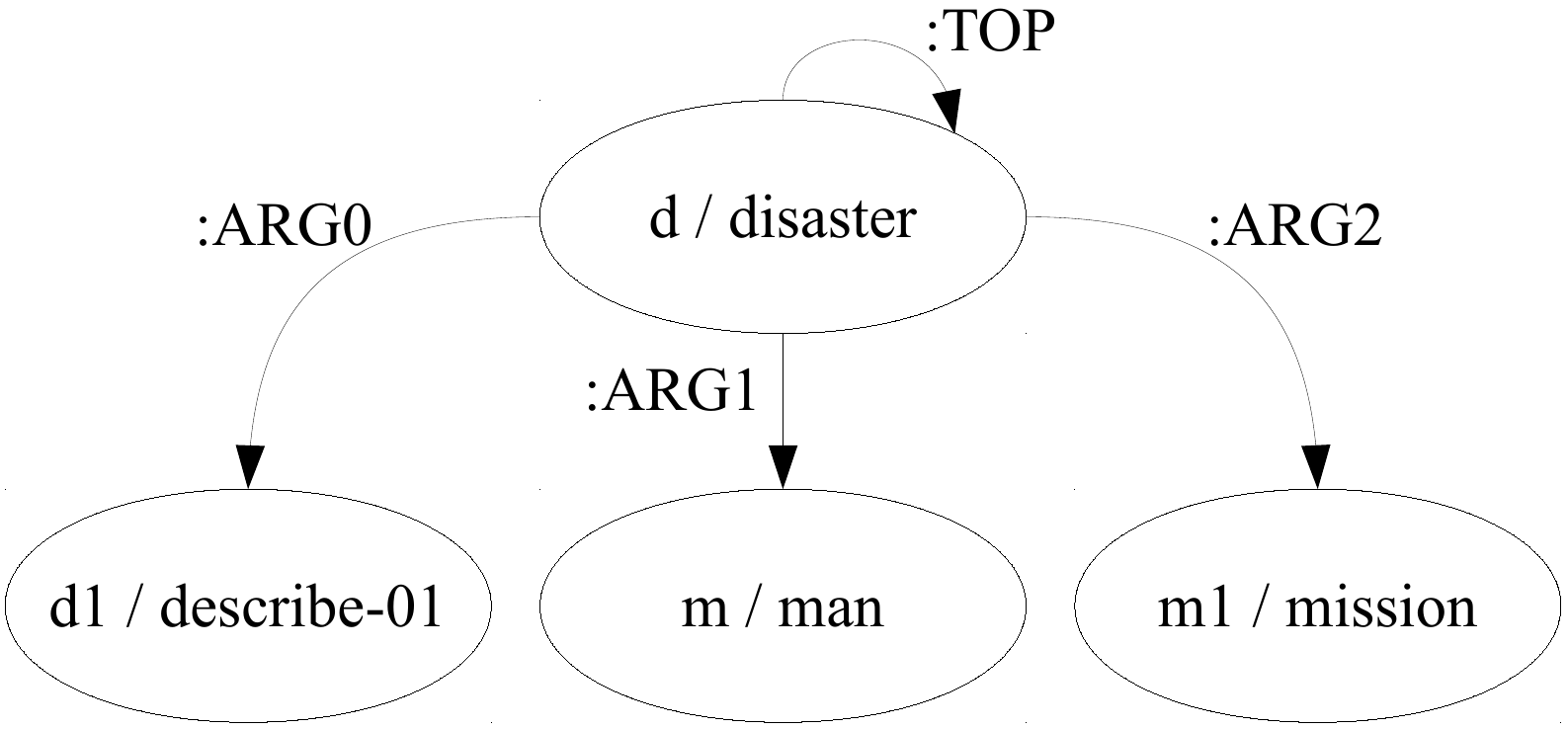}
		\caption{Test AMR}
		\label{fig:test}
	\end{minipage}
\end{figure}

For example, when evaluating the AMR graph in Fig. \ref{fig:test} against the AMR in Fig. \ref{fig:gold}, 
\textit{smatch} returns {\tt M} equal to four ({\tt disaster}, {\tt describe-01}, 
{\tt man}, and {\tt mission}), {\tt C} equal to eight ({\tt disaster}, {\tt describe-01}, {\tt man}, 
{\tt mission}, {\tt TOP}, {\tt ARG0}, {\tt ARG1}, and {\tt ARG2}), and {\tt T} equal to eight. 
So, precision and recall are equal to $4/8 = 0.5$. 

As we may see, \textit{smatch} adds a \textit{TOP} relation in the structure. This self-relation is not 
provided by AMR language and it distorts the analysis because if a parser correctly identifies the root 
of the graph, \textit{smatch} will compute the root concept and the \textit{TOP} relation as correct, 
otherwise, it will compute only the root concept as correct. In addition, \textit{smatch} is not 
considering the dependence of the elements. The other issues will be detailed later.

Thereby, we believe that assessing the dependence in which the elements are arranged in the AMR structure 
may help to better understand the semantic analyzers potentialities and limitations and to produce better 
applications.

Given these shortcomings and inspired by Damonte et al. \cite{damonte2017parser}  to better understand the 
limitations of AMR parsers and to find their strong points, we propose a new metric for evaluating AMR 
parsers, named SEMA (Semantic Evaluation Metric for AMR). Our metric deals with these issues of the 
\textit{smatch} metric, presenting a new way to evaluate concepts and relations in AMR structures, 
computing precision, recall, and f-score values between two AMRs. Moreover, we compare \textit{smatch} and 
SEMA, using four well-known AMR parsers in order to analyze the differences between the metrics and, finally, 
we discuss the obtained results.

In what follows, Sect.~\ref{sec:related} presents the essential related work. In Sect.~\ref{sec:funda}, 
we introduce the AMR fundamentals. Sect.~\ref{sec:metric} details our developed metric. 
In Sect.~\ref{sec:compa}, we compare \textit{smatch} and SEMA and, finally, Sect.~\ref{sec:final} 
concludes the paper.

\section{Related Work} \label{sec:related}

Compared to traditional meaning representations, AMR is a relatively new representation, as well as AMR 
parsing is a new task. Thus, there are few works involving semantic representation measurements.

Allen et al. \cite{allen2008deep} adopted a logical form representation for evaluating 
its semantic representation. The authors proposed a metric that computes the maximum score by any 
alignment among logical form graphs. This representation needs an alignment between the input sentences 
and the semantic analysis. However, the authors did not address how to determine the alignments.

Dridan and Oepen \cite{dridan2011parser} directly evaluated a semantic parser output by 
comparing semantic sub-structures. The authors also adopted a logical form representation for evaluating its 
semantic representation. For that, the authors required an alignment between sentence spans and semantic 
sub-structures. One limitation of that metric is the need for an alignment between the input sentences and 
their semantic analyses.

Cai and Knight \cite{cai2013smatch} developed a metric named \textit{smatch} that calculates 
the degree of overlap between two AMR structures. The metric computes the maximum f-score obtainable via 
one-to-one matching of variables between two AMRs.

As the \textit{smatch} metric, our metric is also focused on AMR structures. However, our metric is more 
robust, because it deals with the several drawbacks that \textit{smatch} has, as the dependence of elements 
(nodes, edges), the self-relation created on the root of the graph, and the weights generated for different 
error types.

\section{AMR Essentials} \label{sec:funda}

Abstract Meaning Representation (AMR) is a semantic representation language designed to capture the meaning 
of a sentence, abstracting away from elements of the surface syntactic structure, such as morphosyntactic 
information and word ordering \cite{banarescu2013abstract}. Hence, words that do not significantly contribute 
to the meaning of a sentence are left out of the annotation.

\begin{figure}[ht]
    \centering
	\frame{\includegraphics[scale=0.41]{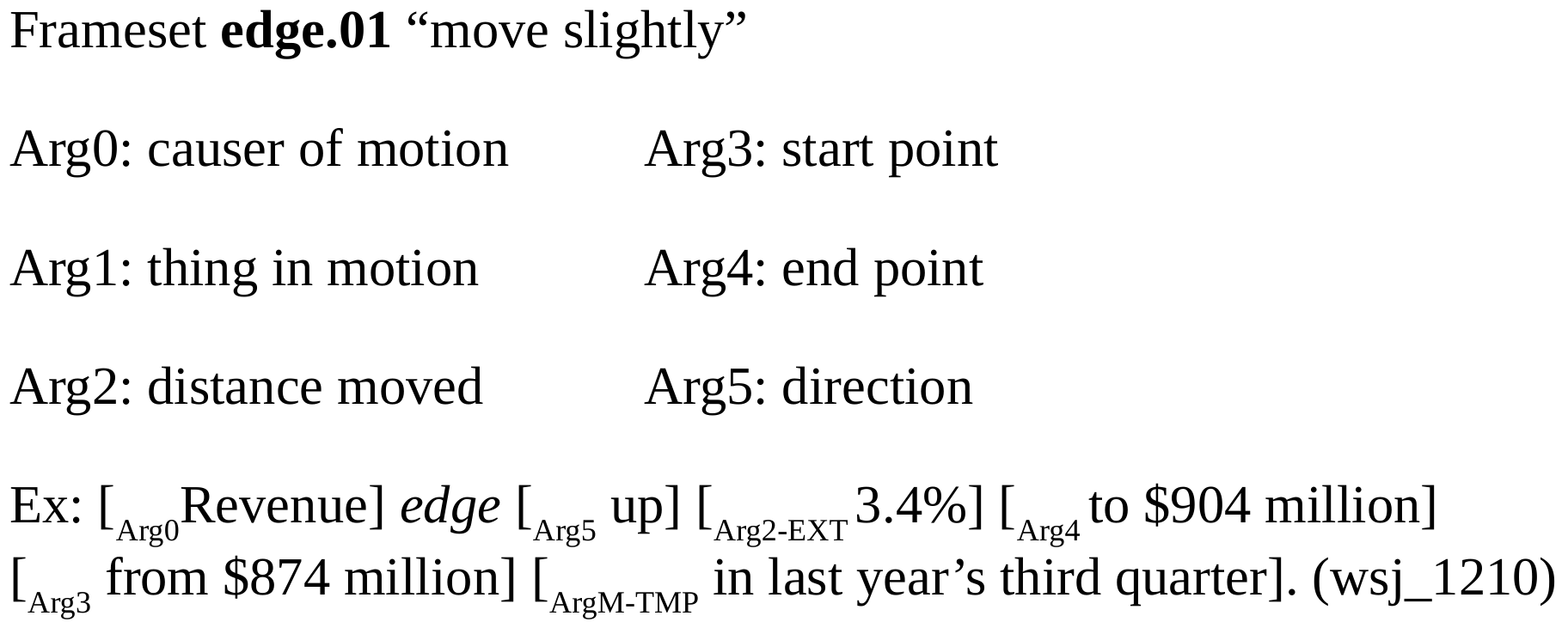}}
	\caption{A PropBank frameset \cite{palmer2005proposition}} \label{fig:frameset}
\end{figure}

AMR focuses on the predicate-argument structure of a sentence, as defined by the PropBank resource 
\cite{palmer2005proposition}. It may be represented as a single-rooted directed acyclic
graph with labeled nodes (concepts) and edges (relations) among them. Nodes represent the main events and 
entities mentioned in a sentence, and edges represent semantic relationships among nodes. AMR concepts are 
either words in their lexicalized forms (e.g., {\tt boy}, {\tt girl}), PropBank framesets ({\tt want-01}, 
{\tt adjust-01}), or special keywords such as {\tt date-entity}, {\tt distance-entity}, 
\\{\tt government-organization}, and others. PropBank framesets are essentially verbs linked to lists of 
possible arguments and their semantic roles. In Fig. \ref{fig:frameset}, we show a PropBank frameset 
example. The frameset {\tt edge.01}, which represents the ``move slightly'' sense, has six 
arguments (Arg 0 to 5).

For semantic relationships, in addition to PropBank semantic roles, AMR adopts approximately 100 additional 
relations. We list some of them below. For more details, we suggest consulting the original paper 
\cite{banarescu2013abstract}.

\begin{description}
	\item[General semantic relations:] :mod, :manner, :location, :name, :polarity
	\item[Relations for quantity:] :quant, :unit, :scale
	\item[Relations for date-entity:] :day, :month, :year, :weekday, :dayperiod
	\item[Relations for list:] :op1, :op2, :op3, and so on
\end{description}

\begin{figure}[h]
	\centering
	\includegraphics[scale=0.40]{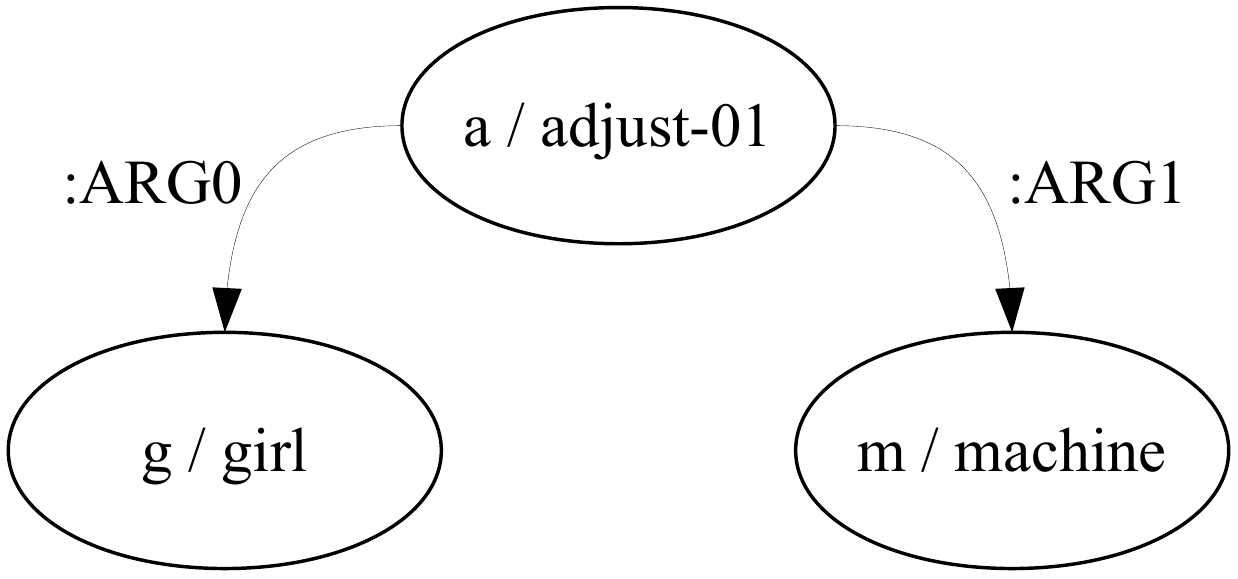}
	\caption{Graph notation} \label{fig:graph}
\end{figure}

In addition to the graph structure, AMR may be represented in two different notations: traditionally, in 
first-order logic; or in the PENMAN notation \cite{matthiessen1991text}, for easier human reading and writing.
For instance, Table \ref{tab:sents} presents sentences with similar senses, which are represented in the 
canonical form in PENMAN format and in the corresponding graph notation, in Figs. \ref{fig:penman} 
and \ref{fig:graph}, respectively.

\begin{minipage}{\textwidth}
    \begin{minipage}[b]{0.5\textwidth}
        \centering
        \captionof{table}{Sentences with similar meaning} \label{tab:sents}
        \begin{tabular}{c}
            \hline
		    \textbf{Sentences}\\
		    \hline
		    The girl made adjustment to the machine.\\
		    The girl adjusted the machine.\\
		    The machine was adjusted by the girl.\\
		    \hline
        \end{tabular}
    \end{minipage}
    \hfill
    \begin{minipage}[t]{0.49\textwidth}
        \centering
        \frame{\includegraphics[scale=0.55]{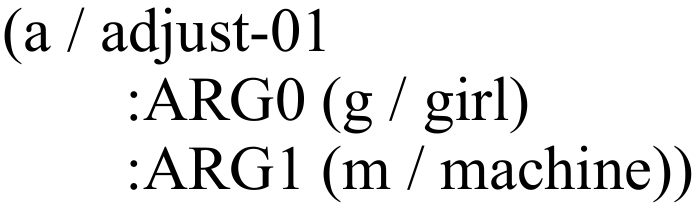}}
	    \captionof{figure}{PENMAN notation} \label{fig:penman}
    \end{minipage}
\end{minipage}

As it is possible to see, AMR assigns the same representation to sentences with the same basic meaning. In 
the example, the concepts are {\tt adjust-01}, {\tt girl}, and {\tt machine} and the relations are 
{\tt :ARG0} and {\tt :ARG1}, represented by labeled directed edges in the graph. In Figs.~\ref{fig:penman} 
and~\ref{fig:graph}, the symbols ``a'', ``g'', and ``m'' are variables and may be re-used in the annotation, 
corresponding to reentrancies (multiple incoming edges) in the graph.

\section{SEMA Metric} \label{sec:metric}

\begin{figure}[h]
    \centering
    \begin{minipage}[b]{0.4\textwidth}
        \centering
        \frame{\includegraphics[scale=0.5]{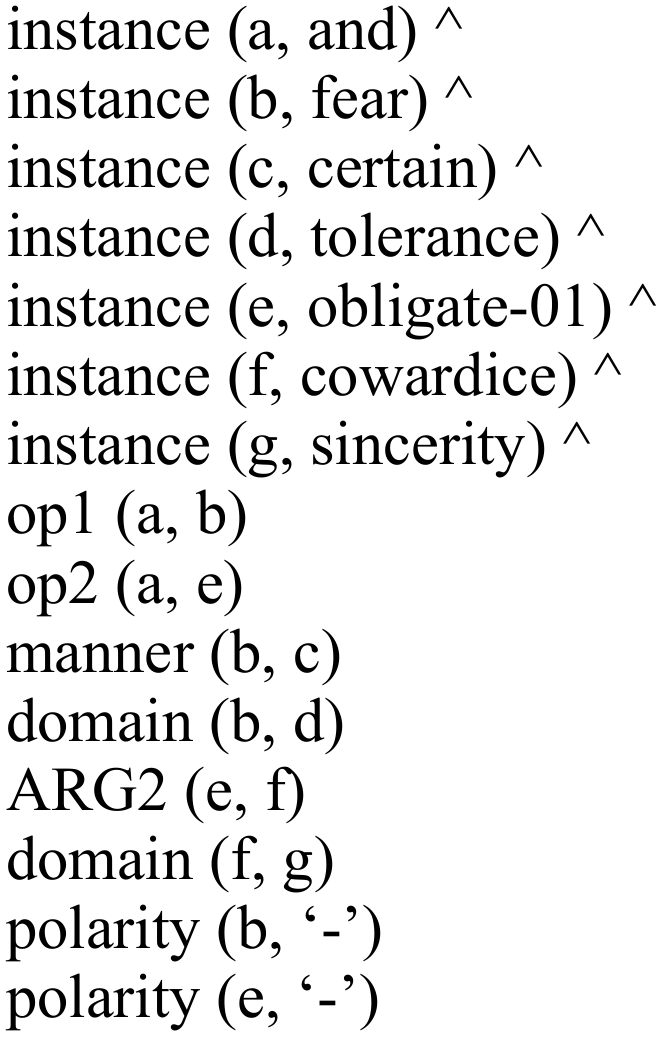}}
        \caption{Reference triples} \label{fig:triples}
    \end{minipage}
    \hfill
    \begin{minipage}[b]{0.59\textwidth}
        \centering
        \includegraphics[scale=0.35]{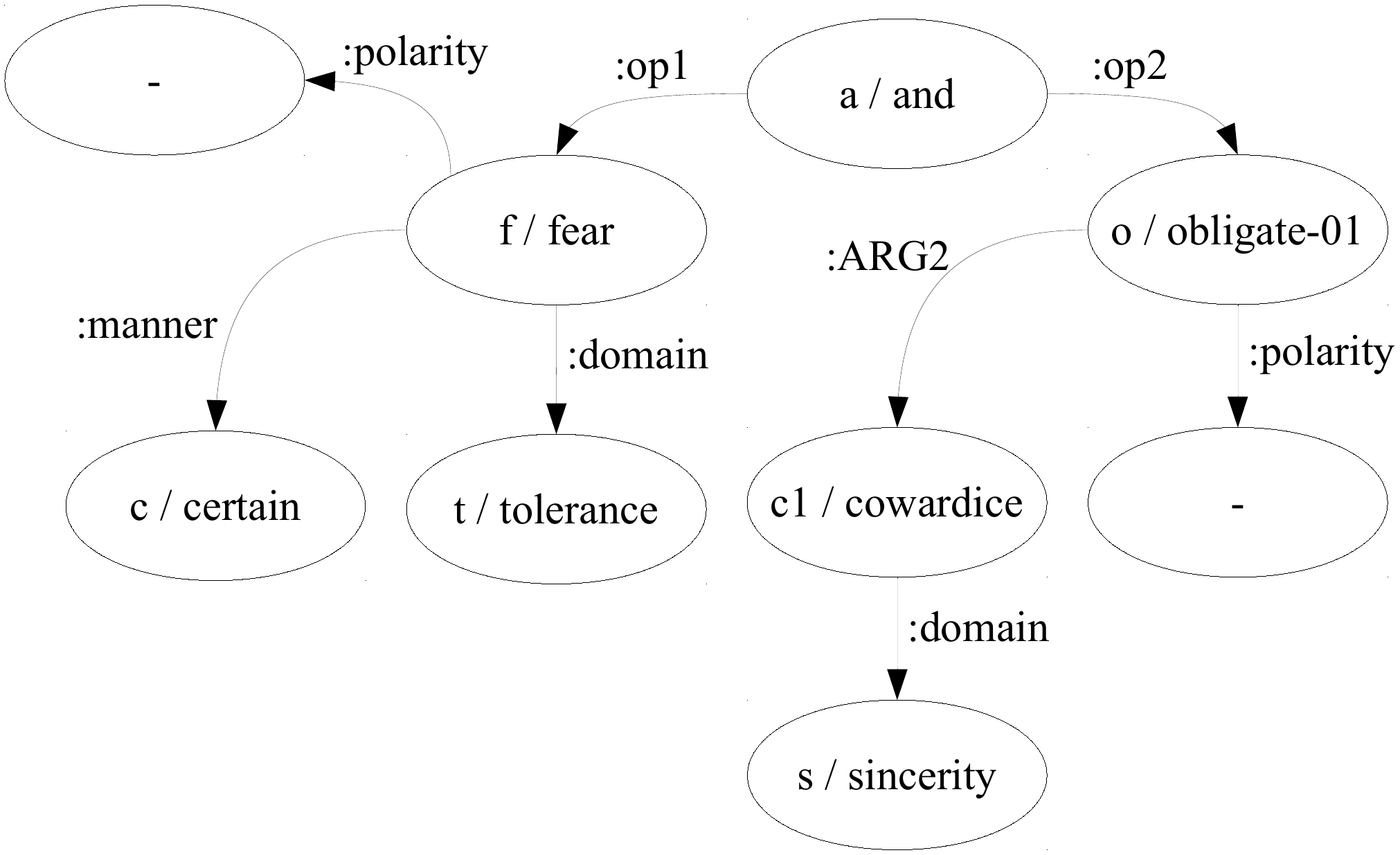}
        \caption{Graph notation for reference triples} \label{fig:amr1}
    \end{minipage}
\end{figure}

Following Cai and Knight \cite{cai2013smatch}, semantic relationships encoded in the AMR 
graph may also be viewed as a conjunction of logical propositions, or triples. For example, suppose that 
the sentence ``Tolerance is certainly not fear, and sincerity does not have to be cowardice.'' produces 
triples according to Fig. \ref{fig:triples} and its graph notation in Fig.~\ref{fig:amr1}.

Each AMR triple takes one of these forms: \textit{relation (variable, concept)}, \textit{relation 
	(variable1, variable2)} or \textit{relation (variable, constant)}. The first form encompasses the 
first seven triples, the second the six triples then, and the third the last two triples in 
Fig. \ref{fig:triples}.

\begin{figure}[h]
    \begin{minipage}[b]{0.4\textwidth}
        \centering
        \frame{\includegraphics[scale=0.5]{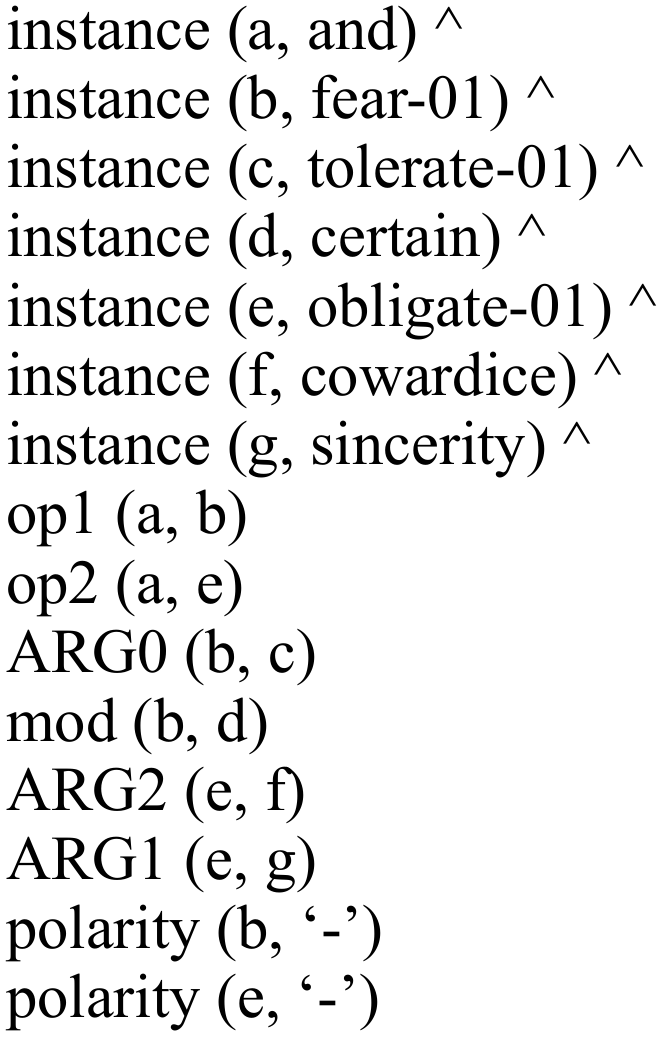}}
        \caption{Test triples} \label{fig:triples2}
    \end{minipage}
    \hfill
    \begin{minipage}[b]{0.59\textwidth}
        \centering
        \includegraphics[scale=0.35]{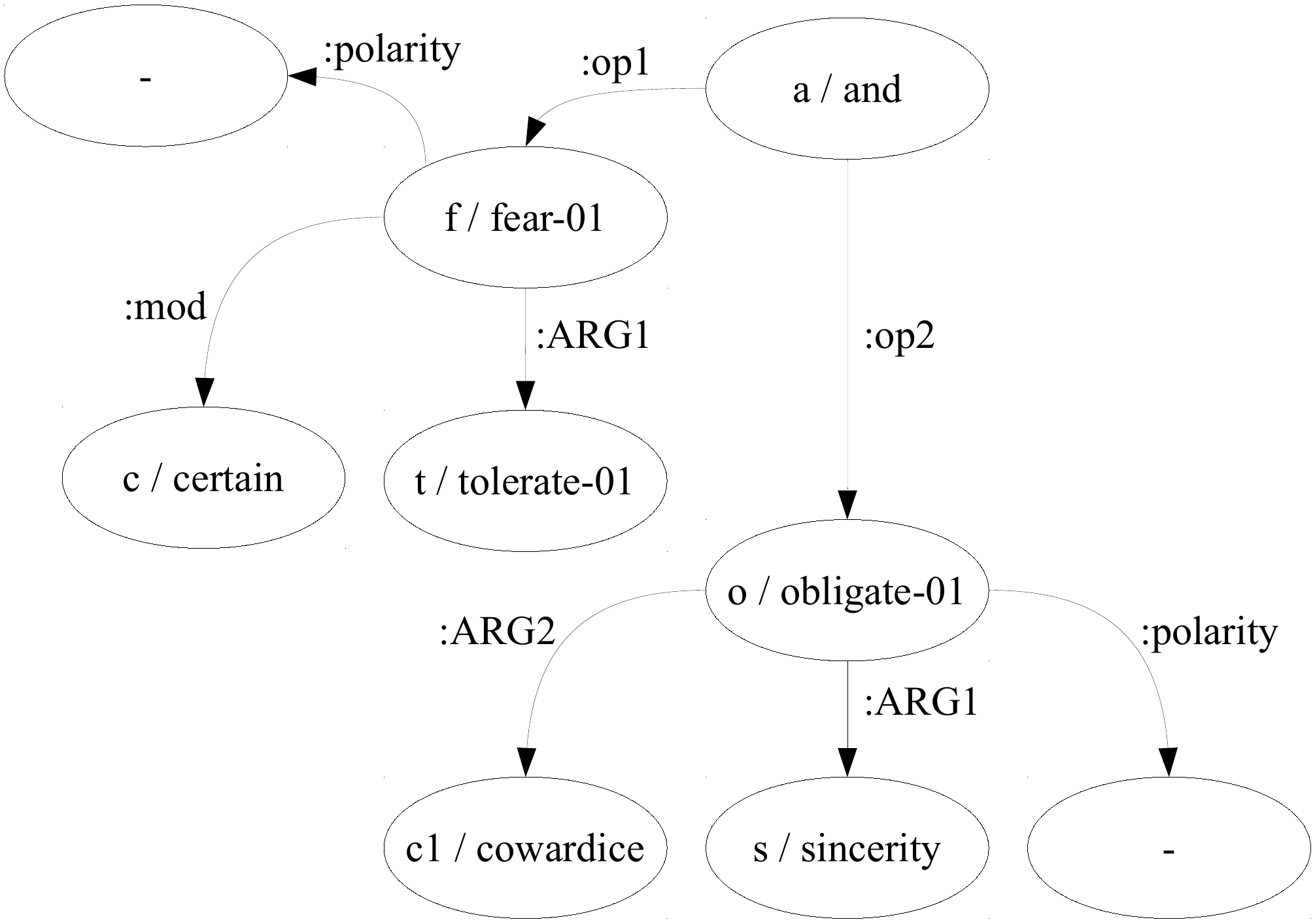}
	    \caption{Graph notation for the test triples} \label{fig:amr2}
    \end{minipage}
\end{figure}

Assuming a second AMR annotation for the same sentence, according to Fig.~\ref{fig:triples2} and 
graphically in Fig. \ref{fig:amr2}, we may compare the two structures considering, for instance, that one 
is produced by a parser and must be compared to the other one, which would be a reference AMR.

Our metric computes precision, recall, and f-score, evaluating the test triples against the reference 
triples, analyzing the root of the graphs and, then, relations and concepts, similar to a Breadth-First 
Search (BFS), taking into account its dependence. 

First, our metric analyzes if the root of the test graph ({\tt and}) belongs to the reference graph, that 
is {\tt and}. We may verify that the two concepts are equal. Thus, the metric computes the concept 
({\tt and}) as correct ({\tt M}), one produced element {\tt and} ({\tt C}), and one reference element 
{\tt and} ({\tt T}). Table \ref{tab:sema_root} presents the root analysis by SEMA.

\begin{table}[!h]
	\setlength{\tabcolsep}{10pt}
	\centering
	\caption{Root analysis} \label{tab:sema_root}
		\begin{tabular}{ccccc}
			\hline
			\textbf{Reference graph} & \textbf{Test graph} & \textbf{M} & \textbf{C} & \textbf{T} \\
			\hline
			{\tt and} & {\tt and} & {\tt and} & {\tt and} & {\tt and} \\
			\hline
		\end{tabular}
\end{table}

Continuing the evaluation, considering the neighbor relations of the root, our metric analyzes if the 
relations {\tt :op1} and {\tt :op2} of the test graph and their parent, which is the root 
of the graph, belong to the reference graph.

Although the two relations are present in reference graph, our metric correctly identifies only the 
{\tt :op2} relation, as the relation {\tt :op1}, in test graph, is connected to the concept {\tt fear-01} 
that is different from the reference graph that is {\tt fear}. In Table~\ref{tab:sema_relations}, we show 
the relations analysis.

\begin{table}[h]
	\setlength{\tabcolsep}{8pt}
	\centering
	\caption{Relations analysis neighbor to the root}
	\label{tab:sema_relations}
	\begin{tabular}{@{}ccccc@{}}
		\hline
		\textbf{Reference graph}                                       & \textbf{Test graph}                                            & \textbf{M} & \textbf{C}                                            & \textbf{T} \\ \hline
		\begin{tabular}[c]{@{}c@{}}{\tt :op1}, {\tt :op2}\end{tabular} & 
		\begin{tabular}[c]{@{}c@{}}{\tt :op1}, {\tt :op2}\end{tabular} & 
		\begin{tabular}[c]{@{}c@{}}{\tt :op2}			 \end{tabular} &  
		\begin{tabular}[c]{@{}c@{}}{\tt :op1}, {\tt :op2}\end{tabular} & 
		\begin{tabular}[c]{@{}c@{}}{\tt :op1}, {\tt :op2}\end{tabular} \\ 
		\hline
	\end{tabular}
\end{table}

After analyzing the relations, our metric analyzes the neighbor concepts of the root, that is, it verifies 
if the concepts {\tt fear-01}, and {\tt obligate-02} of the test graph and their parent, which is the root 
of the graph, belong to the reference graph. 

\begin{table}[h]
    \centering
	\caption{Concepts analysis}
	\label{tab:sema_concepts}
		\begin{tabular}{@{}ccccc@{}}
			\hline
			\textbf{Reference graph} & \textbf{Test graph} & \textbf{M} & \textbf{C} & \textbf{T} \\ 
			\hline
			\begin{tabular}[c]{@{}c@{}}{\tt fear}, \\ {\tt obligate-01}		\end{tabular} & \begin{tabular}[c]{@{}c@{}}{\tt fear-01}, \\ {\tt obligate-01}	\end{tabular} & \begin{tabular}[c]{@{}c@{}}{\tt obligate-01}					\end{tabular} & \begin{tabular}[c]{@{}c@{}}{\tt fear-01}, \\ {\tt obligate-01}	\end{tabular} & 
			\begin{tabular}[c]{@{}c@{}}{\tt fear}, \\ {\tt obligate-01}		\end{tabular} \\ 
			\hline
		\end{tabular}%
\end{table}

As one may see, the concept {\tt obligate-01} is correct and the concept {\tt fear-01} is wrong, since the 
correct concept is {\tt fear}. So, the metric computes correctly one element {\tt fear}, shown in 
Table \ref{tab:sema_concepts}.

In the same manner, our metric will calculate the remaining relations and concepts. At the end of the 
evaluation, our metric returns six correct triples \{\textit{instance (a, and), 
instance (e, obligate-01), op2 (a, e), instance (f, cowardice), ARG2 (e, f), polarity (e, `-')}\} and both
test and reference AMRs produced fifteen triples. So, precision, recall, and f-score are equal to 
$6/15 = 0.40$, respectively. 

\begin{figure}[ht]
	\centering
	\includegraphics[scale=0.4]{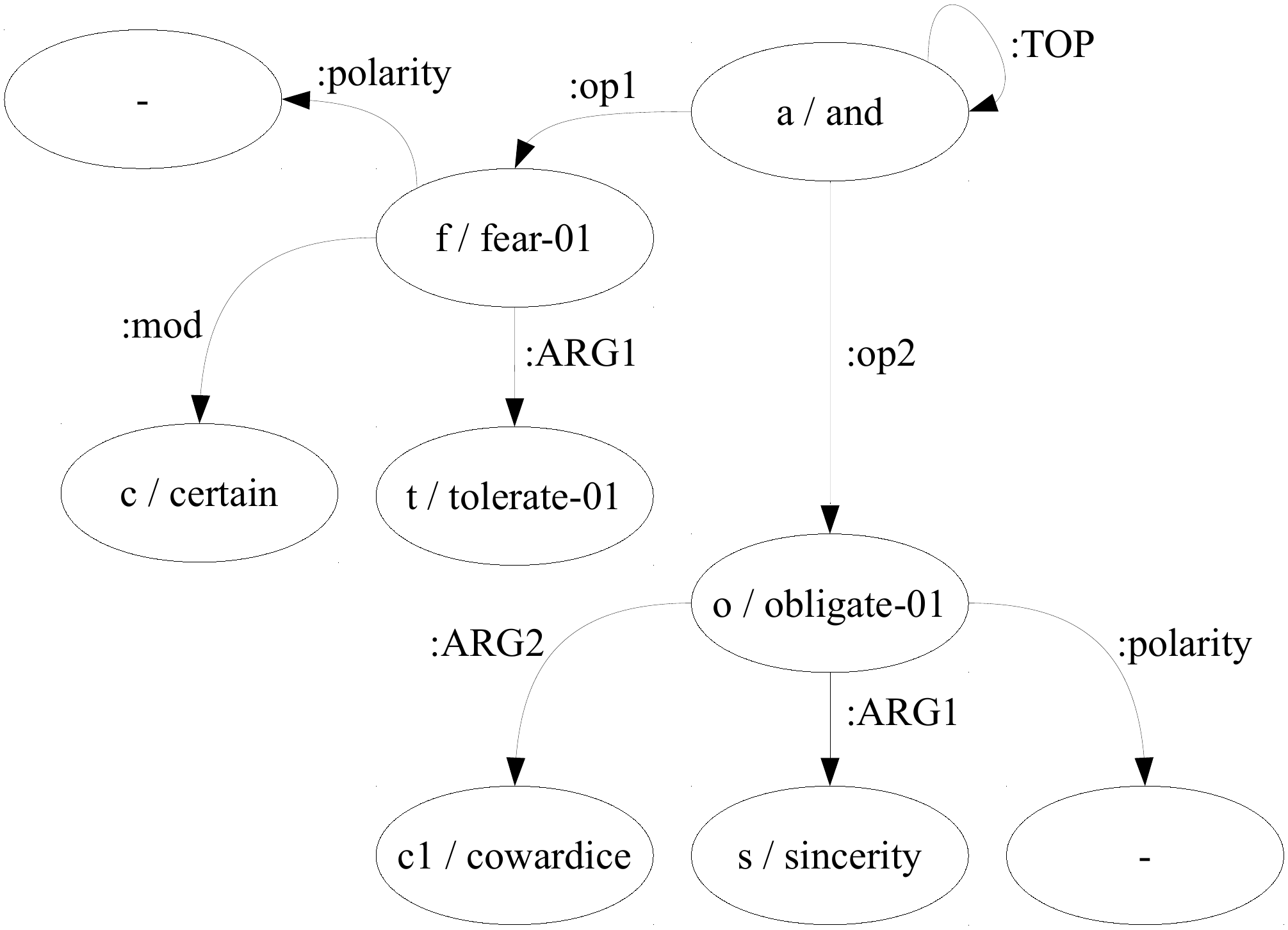}
	\caption{AMR considered by \textit{smatch} metric} \label{fig:smatch}
\end{figure}

Analyzing the previous example, the \textit{smatch} metric returns as precision, recall, and 
f-score values equal to $0.69$ for each measure. \textit{Smatch} considers as correct the triples 
\{\textit{instance (a, and), instance (e, obligate-01), instance (d, certain), instance (g, sincerity), 
instance (f, cowardice), op1 (a, b), op2 (a, e) ARG2 (e, f), polarity (b, `-'), polarity (e, `-'), 
TOP (a, `and')}\} . The metric tries to maximize the f-score, so, it does not evaluate the dependence of 
the elements in the AMR structure. Besides that, the \textit{smatch} scores the root {\tt and} and its 
self-relation {\tt :TOP}, distorting the analysis\footnote{The result may be confirmed at 
	\url{https://amr.isi.edu/eval/smatch/compare.html}. We also checked the available source code 
	\url{https://github.com/snowblink14/smatch}} (see Fig. \ref{fig:smatch}).

In contrast to the \textit{smatch} metric, our metric considers the dependence of the elements arranged 
on a graph, i.e., the metric evaluates the relations/concepts and their parents. Furthermore, our metric 
does not create a {\tt :TOP} relation at the root of the graph, not distorting the evaluation and making the 
analysis fairer than \textit{smatch} metric. More than that, our metric produces a deterministic result 
since it works as a Breadth-first search where in the worst-case the performance is $O(|V| + |E|)$, which 
is faster than to compute the maximum score via one-to-one matching of variable, as the \textit{smatch} metric.

\begin{figure}[h]
	\centering
	\includegraphics[scale=0.45]{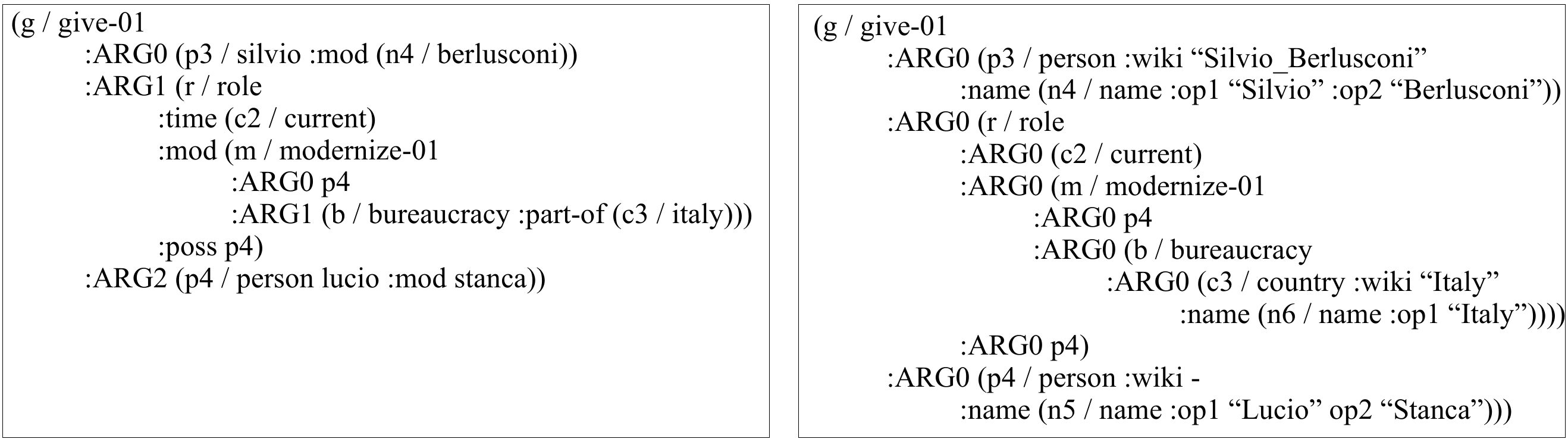}
	\caption{Sentence ``Silvio Berlusconi gave Lucio Stanca his current role of modernizing Italy's 
		bureaucracy'' parsed by two parsers \cite{damonte2017parser}} \label{fig:error_types}
\end{figure}

In addition to the above shortcomings, Damonte and Cohen \cite{damonte2017parser}  detected that 
\textit{smatch} has weights for different error types. For example, considering two parses for the 
sentence ``Silvio Berlusconi gave Lucio Stanca his current role of modernizing Italy's bureaucracy'', 
in Fig. \ref{fig:error_types}.

At the left, the output of a parser (\textit{Parser 1}) is not able to deal with named entities. At the right, 
in the output of other parser (\textit{Parser 2}), except for {\tt :name}, {\tt :op}, and {\tt :wiki} 
the relation label {\tt :ARG0} is always used. The \textit{smatch} scores for two parses are $0.56$ and 
$0.78$ for f-score, respectively. Despite both parses make obvious mistakes, three named entity errors 
in \textit{Parse 1} are considered more important than six wrong labels in \textit{Parse 2}, according to 
Damonte et al. \cite{damonte2017parser}. SEMA metric solves that issue by assigning equal weights to all 
relations, making the evaluation more robust than \textit{smatch}.

By analyzing AMRs according to SEMA, we may measure precision, recall, and f-score for instance and 
relation identification tasks, and thus, understand better the AMR parsing task due to a more 
fine-grained analysis. A demo version and the source code of our metric is available at 
\url{https://github.com/rafaelanchieta/sema}. In what follows, we compared our metric with \textit{smatch} 
using four well-known AMR parsers.

\section{Evaluation} \label{sec:compa}

In order to compare our metric with \textit{Smatch}, we chose four AMR parsers for English: JAMR parser 
\cite{flanigan2014discriminative}, AMREager parser \cite{damonte2017parser}, Neural AMR Parser 
\cite{van2017neural}, and AMR Graph Prediction Parser \cite{chunchuan-amrgraph}. These parsers were chosen 
because they handle the parsing task differently and they are publicly available.

We focused on two datasets: LDC2015E86~(R1), which consists of $16,833$, $1,368$, and $1,371$ sentences in 
training, development, and testing sets, respectively, and LDC2016E25 (R2), which contains $36,521$ training 
sentences, and the same sentences for development and testing as R1. 
Table~\ref{tab:comparison} shows the comparison between the SEMA and \textit{smatch} metrics on the test set.

\begin{table}[h]
	\setlength{\tabcolsep}{6pt}
    \centering
	\caption{Comparison between SEMA and Smatch metrics on the test set}
	\label{tab:comparison}
		\begin{tabular}{@{}cclll|lll@{}}
			\hline
			\multirow{2}{*}{\textbf{Parser}} & \multirow{2}{*}{\textbf{Train. Data}} & \multicolumn{3}{c}{\textbf{SEMA}} & \multicolumn{3}{c}{\textbf{Smatch}} \\ \cline{3-8}
			&                                & \multicolumn{1}{c}{P} & \multicolumn{1}{c}{R} & \multicolumn{1}{c}{F} & \multicolumn{1}{c}{P} & \multicolumn{1}{c}{R} & \multicolumn{1}{c}{F} \\ 
			\hline
			JAMR 		& R1 & 0.61      & 0.57      & 0.59      & 0.70       & 0.64       & 0.67      \\
			AMREAger    & R1 & 0.59      & 0.54      & 0.56      & 0.67       & 0.62       & 0.64      \\
			Neural AMR  & R2 & 0.67      & 0.59      & 0.63      & 0.76       & 0.67       & 0.71      \\
			AMR Graph P.& R2 & 0.67      & 0.64      & 0.66      & 0.75       & 0.72       & 0.74      \\
			\hline
		\end{tabular}
\end{table}

As shown in Table \ref{tab:comparison}, our metric is stricter than \textit{smatch} metric. In order to 
understand these values and how the metrics deal with graphs of different sizes, we carried out a detailed 
evaluation.

We calculated the average number of relations in the test set and found that each sentence has $19.8$ 
relations on average. Thus, we organized the test set into two sets: those sentences with number of 
relations below the average ($799$ sentences) and those with number of relations above the average 
($572$ sentences) and compared the SEMA and \textit{smatch} metrics. Tables~\ref{tab:below_average} 
and \ref{tab:above_average} present the results.


As shown in Tables \ref{tab:below_average} and \ref{tab:above_average}, in both configurations 
\textit{smatch} values were superior to SEMA values.
This is due to two main factors: 


\begin{enumerate}
	\item The distorted analysis of the relation \textit{TOP};
	\item A large number of concepts and relations not properly evaluated by \textit{smatch}.
\end{enumerate}

In the first factor, in 44.75\% of the number of relations below the average and in 77.5\% of the number 
of relations above the average, the parsers did not correctly produce the root of the graph, and, 
even so, \textit{smatch} considered the roots as correct because the concepts were present in the graph. 

\begin{table}[h]
	\centering
	\begin{minipage}[b]{0.49\textwidth}
		\caption{For number of relation below the average}
		\label{tab:below_average}
		\resizebox{\textwidth}{!}{%
			\begin{tabular}{@{}cclll|lll@{}}
				\hline
				\multirow{2}{*}{\textbf{Parser}} & \multirow{2}{*}{\textbf{Train. Data}} & \multicolumn{3}{c}{\textbf{SEMA}} & \multicolumn{3}{c}{\textbf{Smatch}} \\ \cline{3-8}
				&                                & \multicolumn{1}{c}{P} & \multicolumn{1}{c}{R} & \multicolumn{1}{c}{F} & \multicolumn{1}{c}{P} & \multicolumn{1}{c}{R} & \multicolumn{1}{c}{F} \\ 
				\hline
				JAMR		& R1 & 0.61      & 0.55      & 0.58      & 0.71       & 0.65       & 0.68      \\
				AMREAger	& R1 & 0.59      & 0.53      & 0.56      & 0.69       & 0.63       & 0.66      \\
				Neural AMR	& R2 & 0.66      & 0.62      & 0.64      & 0.76       & 0.72       & 0.74      \\
				AMR Graph P.& R2 & 0.66      & 0.64      & 0.65      & 0.75       & 0.73       & 0.74      \\
				\hline
			\end{tabular}%
		}
	\end{minipage}
	\hfill
	\begin{minipage}[b]{0.49\textwidth}
		\caption{For Number of relations above the average}
		\label{tab:above_average}
		\resizebox{\textwidth}{!}{%
			\begin{tabular}{@{}cclll|lll@{}}
				\hline
				\multirow{2}{*}{\textbf{Parser}} & \multirow{2}{*}{\textbf{Train. Data}} & \multicolumn{3}{c}{\textbf{SEMA}} & \multicolumn{3}{c}{\textbf{Smatch}} \\ \cline{3-8}
				&                                & \multicolumn{1}{c}{P} & \multicolumn{1}{c}{R} & \multicolumn{1}{c}{F} & \multicolumn{1}{c}{P} & \multicolumn{1}{c}{R} & \multicolumn{1}{c}{F} \\ 
				\hline
				JAMR		& R1 & 0.62      & 0.58      & 0.60      & 0.69       & 0.64       & 0.66      \\
				AMREAger    & R1 & 0.58      & 0.54      & 0.56      & 0.66       & 0.61       & 0.63      \\
				Neural AMR  & R2 & 0.68      & 0.57      & 0.62      & 0.74       & 0.63       & 0.68      \\
				AMR Graph P.& R2 & 0.68      & 0.65      & 0.67      & 0.75       & 0.72       & 0.73      \\
				\hline
			\end{tabular}%
		}
	\end{minipage}
\end{table}

\begin{figure}[h]
	\begin{minipage}[b]{0.4\textwidth}
		\centering
		\includegraphics[scale=0.3]{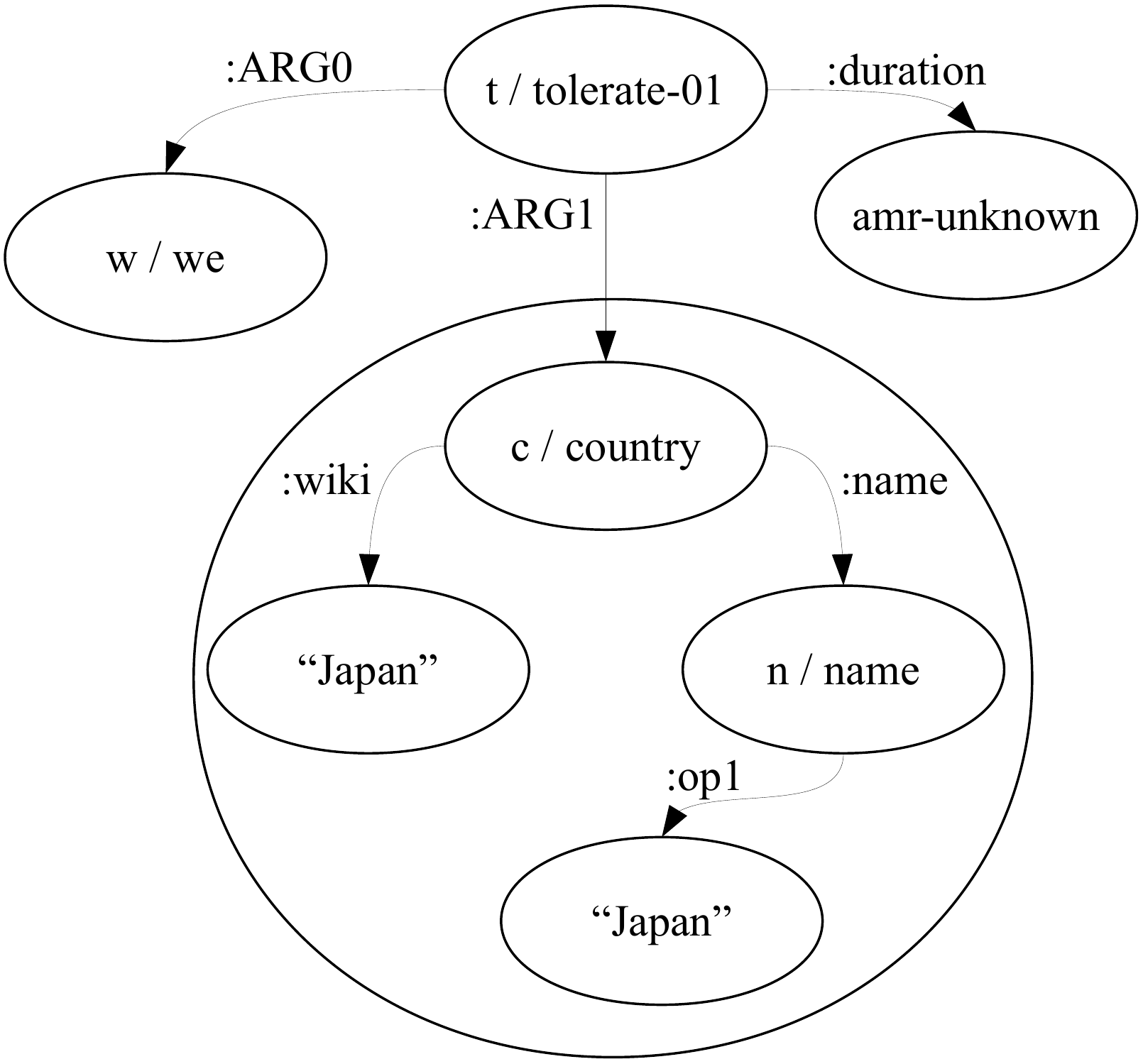}
		\caption{Reference AMR graph} \label{fig:subgraph-gold}
	\end{minipage}
	\hfill
	\begin{minipage}[b]{0.52\textwidth}
		\centering
		\includegraphics[scale=0.3]{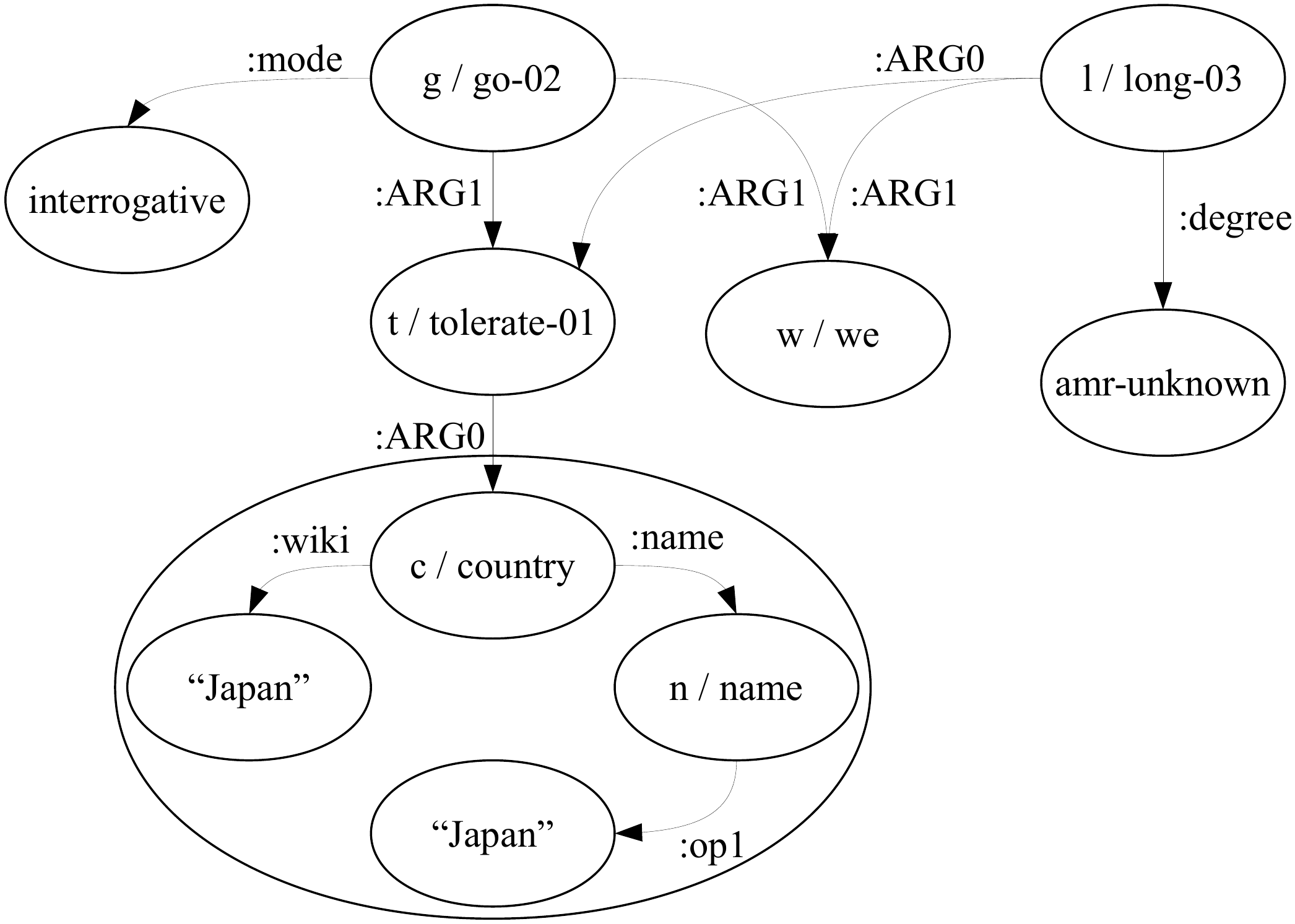}
		\caption{AMR graph generated by a parser} \label{fig:subgraph-test}
	\end{minipage}
\end{figure}

For the second factor, consider the sentence ``How long are we going to tolerate Japan?'', which was manually 
annotated as in Fig.~\ref{fig:subgraph-gold}. The AMR graph has six relations and seven concepts 
(11 triples). For the same sentence, an AMR parser generated the AMR graph in 
Fig.~\ref{fig:subgraph-test}, which has ten relations and concepts~(17 triples).

We may see that the AMR parser produced a subgraph similar to a subgraph that was manually annotated. 
Despite there are other concepts in the AMR graph produced by the parser that are present in reference AMR 
graph, as: {\tt tolerate-01}, {\tt we} and {\tt amr-unknown}, their dependents and/or relations are wrong. 
Hence, the SEMA metric considers these concepts as wrong. For instance, the concept {\tt tolerate-01}, in 
the reference AMR graph, is the root of the graph, whereas, in the AMR produced by the parser, the root is 
the concept {\tt go-02}. The root {\tt go-02} is connected to the concept {\tt tolerate-01} through the 
{\tt :ARG1} relation. Finally, the concept {\tt tolerate-01} in both graphs is connected to the concept 
{\tt country} but by different relations: {\tt :ARG0} and {\tt :ARG1} for the AMR generated by the parser 
and reference AMR graph, respectively.

Due to these distinctions, our metric evaluates the connection with the subgraph as wrong since its relation
is different from the reference AMR graph. On the other hand, the \textit{smatch} metric evaluates as 
correct the concepts {\tt we} and {\tt amr-unknown}, although they are not connected to the concept 
{\tt tolerate-01}. 
Thus, the \textit{smatch} returns $0.44$, $0.67$, and $0.53$, while the SEMA returns $0.29$, $0.45$, and 
$0.36$, for precision, recall, and f-score, respectively.

Even though our metric is stricter than \textit{smatch} metric, we believe that SEMA is fairer and more 
robust than \textit{smatch}. As AMR parsing task is on the semantic level, the dependence of the elements 
in AMR structure should be analyzed. More than that, SEMA metrics neither creates a TOP self-relation on the 
root of the graph nor assigns weights for different error types, not distorting the analysis. In the way \textit{smatch} is currently computed, several parsing problems are overlooked.

\section{Final Remarks} \label{sec:final}

In this paper, we presented a new metric for evaluating AMR structures. This metric analyzes the 
dependence in which the elements are arranged in the AMR structure and deals with other shortcomings of the 
\textit{smatch} metric, as a self-relation produced on the root of the graph, which distorts the analysis, 
and weights for different error types.
We compared our metric with the \textit{smatch} metric, using four AMR parser and showed that, in general, 
our metric is stricter than \textit{smatch} metric. However, we believe that our metric is fairer and 
robust than \textit{smatch} since several parsing problems are being overlooked by \textit{smatch}. 
In addition, we also showed that for both small and large graphs, the parsers have difficulty in learning 
the dependence of the elements, and even so, \textit{smatch} considers as correct several elements.

As future work, we intend to investigate how to adapt our metric to other semantic representations.

\section*{Acknowledgments}

The authors are grateful to FAPESP and IFPI for supporting this work.

%
%
%
\bibliographystyle{splncs04}
\bibliography{reference}
\end{document}